\documentclass[letterpaper, 10 pt, conference]{ieeeconf}  

\IEEEoverridecommandlockouts                              

\usepackage{graphics} 
\usepackage{epsfig} 
\usepackage{amsmath} 
\usepackage{amssymb}  

\DeclareMathOperator*{\argmax}{argmax}
\usepackage{latexsym}
\usepackage{graphicx}
\usepackage{amsfonts,amssymb,amsmath}
\usepackage{hyperref}
\usepackage[T1]{fontenc}
\usepackage{cite}
\usepackage{tikz}
\usepackage[nogroupskip,acronyms,nopostdot,style=super,nonumberlist,toc]{glossaries}
\usepackage{tabularx}
\usepackage{xfrac}
\usepackage{pgfplots}
\pgfplotsset{compat=1.16}
\usepgfplotslibrary{groupplots}
\usepackage[caption=false,font=footnotesize]{subfig}
\usepackage{siunitx}
\newacronym{ai}{AI}{artificial intelligence}%
\newacronym{pcam}{PCAM}{pedestrian crash avoidance mitigation}%
\newacronym{drl}{DRL}{deep reinforcement learning}
\newacronym{dl}{DL}{deep learning}
\newacronym{ml}{ML}{machine learning}
\newacronym{rl}{RL}{reinforcement learning}
\newacronym{ad}{AD}{autonomous driving}
\newacronym{av}{AV}{autonomous vehicle}
\newacronym{dnn}{DNN}{deep neural network}
\newacronym{ann}{ANN}{artificial neural network}
\newacronym{nn}{NN}{neural network}
\newacronym{dqn}{DQN}{deep Q-network}
\newacronym{rnn}{RNN}{recurrent neural network}
\newacronym{rdqn}{RDQN}{recurrent deep Q-network}
\newacronym{ddqn}{DDQN}{double deep Q-network}
\newacronym{marl}{MARL}{multi-agent reinforcement learning}
\newacronym{dmarl}{DMARL}{deep multi-agent reinforcement learning}
\newacronym{mdp}{MDP}{Markov decision process}
\newacronym{mlp}{MLP}{multilayer perceptron}
\newacronym{mpc}{MPC}{model predictive control}
\newacronym{its}{ITS}{intelligent transportation systems}
\newacronym{ttc}{TTC}{time-to-collision}
\newacronym{ddpg}{DDPG}{deep deterministic policy gradient}
\newacronym{vae}{VAE}{variational auto-encoder}
\newacronym{mas}{MAS}{multi-agent system}
\newacronym{mal}{MAL}{multi-agent learning}
\newacronym{cad}{CAD}{connected autonomous driving}
\newacronym{mc}{MC}{Monte Carlo}
\newacronym{dp}{DP}{dynamic programming}
\newacronym{td}{TD}{temporal-difference}
\newacronym{sgd}{SGD}{stochastic gradient descent}
\newacronym{mse}{MSE}{mean squared error}
\newacronym{per}{PER}{prioritized experience replay}
\newacronym{a2c}{A2C}{advantage actor critic}
\newacronym{sg}{SG}{stochastic game}
\newacronym{mg}{MG}{Markov game}
\newacronym{br}{BR}{best response}
\newacronym{pomdp}{POMDP}{partially observable Markov decision process}
\newacronym{pomg}{POMG}{partially observable Markov game}
\newacronym{iql}{IQL}{independent Q-learning}
\newacronym{idqn}{IDQN}{independent deep Q-network}
\newacronym{dpomdp}{dec-POMDP}{decentralized partially observable Markov decision process}
\newacronym{nrmse}{NRMSE}{normalized root-mean-square error}
\newacronym{rmse}{RMSE}{root-mean-square error}
\newacronym{snr}{SNR}{signal-to-noise ratio}
\newacronym{cos}{COS}{coordinate system}

\title{\LARGE \bf
Modeling Interactions of Autonomous Vehicles and Pedestrians with Deep Multi-Agent Reinforcement Learning for Collision Avoidance
}

\author{Raphael~Trumpp$^{1}$,
        Harald~Bayerlein$^{1}$,
        and~David~Gesbert$^{2}$
\thanks{R. Trumpp and H. Bayerlein were supported by the Chair of Cyber-Physical Systems in Production Engineering at TUM. H. Bayerlein and D. Gesbert were partially supported by the French government, through the 3IA Côte d’Azur project number ANR-19-P3IA-0002, as well as by the TSN CARNOT Institute under project Robots4IoT.}
\thanks{$^{1}$R. Trumpp and H. Bayerlein are with the TUM School of Engineering and Design, Technical University of Munich, Germany, e-mail: {\{raphael.trumpp, h.bayerlein\}@tum.de}.}
\thanks{$^{2}$D. Gesbert is with the Communication Systems Department, \mbox{EURECOM}, Sophia Antipolis, France, e-mail: david.gesbert@eurecom.fr}.}%

\begin{document}

\maketitle
\thispagestyle{empty}
\pagestyle{empty}

\begin{abstract}
Reliable \gls*{pcam} systems are crucial components of safe \glspl*{av}. The nature of the vehicle-pedestrian interaction where decisions of one agent directly affect the other agent's optimal behavior, and vice versa, is a challenging yet often neglected aspect of such systems. We address this issue by modeling a \gls*{mdp} for a simulated \gls*{av}-pedestrian interaction at an unmarked crosswalk. The \gls*{av}'s \gls*{pcam} decision policy is learned through \gls*{drl}. Since modeling pedestrians realistically is challenging, we compare two levels of intelligent pedestrian behavior. While the baseline model follows a predefined strategy, our advanced pedestrian model is defined as a second \gls*{drl} agent. This model captures continuous learning and the uncertainty inherent in human behavior, making the \gls*{av}-pedestrian interaction a \gls*{dmarl} problem. We benchmark the developed \gls*{pcam} systems according to the collision rate and the resulting traffic flow efficiency with a focus on the influence of observation uncertainty on the decision-making of the agents. The results show that the \gls*{av} is able to completely mitigate collisions under the majority of the investigated conditions and that the \gls*{drl} pedestrian model learns an intelligent crossing behavior.%
\end{abstract}

\glsresetall

\IEEEpeerreviewmaketitle

\section{Introduction}
While the advent of modern \gls*{ai}-based methods holds promise to solve many problems in \gls*{ad}, e.g., the perception of the vehicle’s environment through \gls*{ai}-based computer vision, decision making in safety-relevant driving situations remains challenging. Particularly critical situations are vehicle-pedestrian interactions where the vehicle is moving forward on a collision path with a pedestrian attempting to cross a street.

According to a study by the Insurance Institute for Highway Safety \cite{jermakian2011primary}, there were around 330,000 crashes involving pedestrians between 2005 and 2009 in the U.S. with 224,000 cases related to situations where the pedestrian was hit by the front of a car; pedestrians were crossing a street in 95\,\% of these accidents. Modern cars are equipped with \gls*{pcam} systems to avoid such collisions, making them a crucial component of future \glspl*{av}. In \cite{schratter2019pedestrian}, a modern \gls*{pcam} system is developed on the basis of reachability analysis in conjunction with a situation-aware trajectory planner. This method necessitates the use of a reliable dynamic model for the movement of the pedestrian posing real-world challenges and neglecting the simultaneous decision-making processes of the \gls*{av} and the pedestrian. \Gls*{drl} offers the possibility to reflect that the \gls*{av}’s action directly influences the pedestrian’s reaction, and vice versa. After successful applications in game-related environments \cite{mnih2015human}, recent research interest is shifting to real-world applications as \gls*{drl} methods offer attractive generalization ability without the need for prior domain information.

The study of Chae et al. \cite{chae2017autonomous} is the first publication in which a \gls*{drl}-based \gls*{pcam} system was developed. Although successfully preventing collisions, the \gls*{av} agent is limited to braking actions which neglects that controlled acceleration can also help to prevent dangerous situations. In a broader sense, Papini et al. \cite{papini2021reinforcement} extend the work of Chae et al. by proposing a \gls*{drl}-based system which restricts an \gls*{av} agent by a learned speed limit. This limit ensures that a collision can always be prevented when a distracted pedestrian decides to cross. In \cite{deshpande2020behavioral}, a grid-based state representation is proposed that allows the \gls*{pcam} system to account for multiple pedestrians simultaneously. While the trained agent is evaluated in CARLA and its advantages are discussed, the system's real-world applicability remains open as the influence of uncertainty, e.g., measurement noise and random pedestrian behavior, is not reflected. We address this challenge in our work by conducting an extensive study on the influence of uncertainty on the agents' performance. The recent work of Deshpande et al. \cite{deshpande2021navigation} introduces multi-objective \gls*{drl} to the interaction of \glspl*{av} with pedestrians but focuses more on the navigation of the \gls*{av} than the pedestrian crossing decision. A general overview of \gls*{drl} in \gls*{ad} can be found in \cite{kiran2021deep}.

Crucially, the mentioned previous works only use simple pedestrian models that raise the question of how realistic the crossing decisions of the pedestrians in the proposed systems are. Behavioral research provides some clues as to what factors influence pedestrians' crossing decisions. In general, the \gls*{ttc} value is a key indicator \cite{rasouli2020autonomous}. A typical limit is a \gls*{ttc} value of less than 3s, which makes it unlikely for pedestrians to attempt a crossing \cite{schmidt2009pedestrians}. While this property is often used to model pedestrian behavior, it is necessary to also consider the social aspects of \gls*{ad}. Millard-Ball \cite{millard2018pedestrians} refers to a situation which he calls \textit{crosswalk chicken}: As pedestrians know that \glspl*{av} will stop if necessary, they perceive a low level of risk and cross more recklessly.

We propose a new perspective on modeling pedestrian crossing behavior by developing an \gls*{av} \gls*{pcam} system through a \gls*{dmarl}-based solution that exploits the continued interaction of two independent, learning agents. In this approach, the \gls*{pcam} policy is optimized while the pedestrian learns to cross the street safely at the same time. Additionally, the following contributions are made in this work:
\begin{itemize}
	\item The proposed \gls*{pcam} system's driving capability is extended beyond similar works with the \gls*{av}'s action space to include braking and acceleration actions; no additional local trajectory planer is needed.
	\item We introduce several pedestrian models of different \textit{intelligence} levels, i.e., we compare \gls*{drl} and \gls*{dmarl} settings to evaluate the influence of a learning pedestrian model on the behavior of the \gls*{av}.
	\item An extensive study on the influence of observation noise on the agents' performance is conducted, and a behavioral analysis shows the robustness of the developed algorithms in the face of uncertainty. 
	\item Our approach is generalized over different scenarios with varying values of the initial \gls*{ttc} value, street width, and pedestrian walking speed.
\end{itemize}
\section{System Model}
The proposed \gls*{pcam} system is developed in a simulated driving scenario of an \gls*{av} facing a single pedestrian at an \textit{unmarked} crosswalk. Note that a large number of crosswalks are unmarked; studies \cite{zegeer2005safety} found no links between increased pedestrian safety and marked crosswalks. There is no priority given to the pedestrian in our scenario and we neglect the presence of other road users. The heterogeneous agents are described as follows: 
\begin{itemize}
    \item AV: Vehicle with fully autonomous driving capabilities (level-5); equipped with high-quality sensors, i.e., measurement noise is reduced to minimal levels. Properties of the \gls*{av} are labeled by superscript $(\cdot)^{\text{AV}}$.  
	\item Pedestrian: Attempts to cross the street from the left or right sidewalk with state estimations of limited reliability accounting for variability in human perception. Superscript $(\cdot)^{\text{ped}}$ marks the pedestrian's variables.
\end{itemize}

\begin{figure}
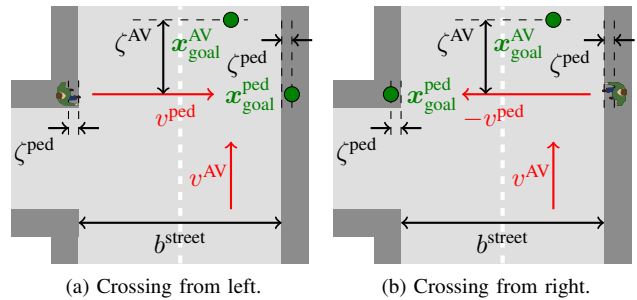

    \centering
    \subfloat[Crossing from left.]{\input{graphics/pedInitLeft.tex}\label{fig_first_case}}
    \hfil
    \subfloat[Crossing from right.]{\input{graphics/pedInitRight.tex}\label{fig_second_case}}
    \caption{Crosswalk geometry with goal positions marked in green and the agents' velocity vectors in red.}
    \label{fig:StreetLayout}
\end{figure}

One AV-pedestrian interaction episode is over after $T \in \mathbb{N}$ time steps, where the time horizon is discretized into equal time slots $t \in [0, T]$ of length $\delta_t$ seconds. We define the simulation in 2D-space, i.e., the \gls*{av}'s position is $\boldsymbol{x}^{\text{AV}}_t=[x^{\text{AV}}_{1,t}, x^{\text{AV}}_{2,t}]^\top \in \mathbb{R}^2$ and $\boldsymbol{x}^{\text{ped}}_t=[x^{\text{ped}}_{1,t}, x^{\text{ped}}_{2,t}]^\top \in \mathbb{R}^2$ is the pedestrian's position, respectively. When an episode starts, the \gls*{av} is facing the crosswalk in front and is positioned at the middle of the right lane of a two-lane street of width $b^{\text{street}}$. Its velocity $v^{\text{AV}}_t \in \mathbb{R}$ is a single component in the vehicle's longitudinal direction. The pedestrian attempts to cross either from the left or the right sidewalk with walking speed $v^{\text{ped}}_t \in \mathbb{R}$; its initial distance to the curb is $\zeta^{\text{ped}}$. When the \gls*{av} has passed the crosswalk by a distance of $\zeta^{\text{AV}}$, the vehicle's goal position $\boldsymbol{x}^{\text{ped}}_{\text{goal}} \in \mathbb{R}^2$ is reached. The pedestrian's episode is over when its position is at a safety distance of $\zeta^{\text{ped}}$ from the street as visualized in \autoref{fig:StreetLayout}.

A collision is defined as the event when the pedestrian, simulated as a point with no dimensions, is inside of the \textit{collision area} of the vehicle which is the \gls*{av}'s dimension plus an additional safety margin $\eta$ around all sides of the \gls*{av}.
The inequality
\begin{equation}
	\label{eqn:collision}
	\eta > |x^{\text{AV}}_{1,t} - x^{\text{ped}}_{1,t}| \wedge \eta > |x^{\text{AV}}_{2,t} - x^{\text{ped}}_{2,t}|,
\end{equation}
is fulfilled in case of a collision. Note that we calculate the \gls*{ttc} value from the \gls*{av}'s center point to the pedestrian's position for simplicity, i.e., a collision can occur at a \gls*{ttc} value marginally larger than zero.

To describe a realistic scenario, measurement noise disturbs the agents' observations according to a multiplicative noise model. Let $s_t$ be the true state signal, then the observation function $\mathcal{O}$ is described by
\begin{equation}
    \label{eqn:ObsFunc}
	\mathcal{O}:z_t =  (1+n_{t}) \cdot s_t,
\end{equation}
with $z_t$ as the disturbed state observation, and $n_{t}$ as the realization of the random noise variable $N$ which follows a Gaussian distribution $N \sim \mathcal{N}(0, \alpha^2)$. As we later investigate several scenarios with different scales of $s_t$, the definition of (\ref{eqn:ObsFunc}) is advantageous since it introduces similar levels of uncertainty among sensor signals of different scale through the single choice of $\alpha$.

As any collision of an \gls*{av} with a pedestrian is considered to be unacceptable in reality, we use the \textit{collision rate} as the main performance measure. Both agents also act self-interested with the motivation to reach their goal position as quickly as possible. This property is reflected by the second performance indicator \textit{traffic flow efficiency}, which describes the average duration of an AV-pedestrian interaction episode. This description is formalized by the utility function
\begin{equation}
	\label{eqn:MainUtility}
	\mathcal{F}^{w} = - T^{w}_{\text{end}}
\end{equation} 
of each agent $w \in \mathcal{W} $ with $\mathcal{W} = \{w^\text{AV}, w^\text{ped}\}$. $T^{\text{AV}}_{\text{end}}$ and $T^{\text{ped}}_{\text{end}}$ describe the time required by the respective agent to reach its goal position. The optimization problem with respect to the trajectory of joint actions $\times_{T}\mathbf{u}_{t}=((u^{\text{AV}}_{0}, u^{\text{ped}}_{0}), \dots, (u^{\text{AV}}_{T}, u^{\text{ped}}_{T}))$ over $T=\max(T^{\text{AV}}_{\text{end}}, T^{\text{ped}}_{\text{end}})$ steps is then given by
\begin{equation}
	\begin{array}{c}
		\max_{\times_{T}\mathbf{u}_{t}} \sum_{w \in \mathcal{W}}  {\mathcal{F}}^{w}. \\
		\text {s.t. } \text{collision} = \text{False}
	\end{array}
\end{equation}
\section{Methodology}
\subsection{Scenarios}
We investigate the effects of modeling the behavior of the street-crossing pedestrian differently by considering the following agents characterized by their level of \textit{intelligence}:
\begin{itemize}
    \item \textit{Level-1} describes \textit{rationally} acting agents. They can adapt their behavior by perceiving their environment continuously but follow a predefined strategy.
    \item Agents of \textit{level-2} can \textit{learn} from their experiences, behave rationally, and explore new strategies. 
\end{itemize}
We use these levels to define several settings for the \gls*{av}'s \gls*{pcam} system using \gls*{drl} and \gls*{dmarl}, respectively. Detailed descriptions of the pedestrian and \gls*{av} models follow in \autoref{subsec:PedestrianModels} and \autoref{subsec:AVModels}.

\subsubsection{Setting-1}
The level-1 pedestrian model is evaluated in conjunction with a learning \gls*{av} agent of level-2 in a \gls*{pomdp} (see \autoref{subsec:POMDB}) system formulation. Most publications (e.g., \cite{deshpande2020behavioral}) up to now have only considered this setting. 

\subsubsection{Setting-2}
Instead of relying on predefined policies, we use a learning agent of level-2 to model the pedestrian. Since the \gls*{av} is also implemented as a \gls*{drl} agent, the system is now a \gls*{mas} modeled as a \gls*{pomg} (see \autoref{subsec:POMG}). To the authors' best knowledge, this is the first approach to use \gls*{dmarl} in the context of vehicle-pedestrian interactions. The proposed \gls*{mas} features \textit{heterogeneous} agents acting in a \textit{semi-cooperative} manner to prevent collisions, but they also aim to fulfill their individual objective of reaching their respective goal position. There is \textit{no direct communication} between agents, but indirect communication via specific actions or behavior signals might be learned. 

\subsubsection{Setting-X}
Another setting is introduced for benchmarking the \gls*{drl} and \gls*{dmarl}-based approaches by defining both the pedestrian and the \gls*{av} as level-1 agents.

\subsection{Pedestrian Models}
\label{subsec:PedestrianModels}
\subsubsection{Level-1}
To resemble a basic but rational human crossing behavior \cite{schmidt2009pedestrians}, we define a pedestrian policy which evaluates the \gls*{ttc} value at each time step $t$ according to
\begin{equation}
	\label{eqn:TTCPedestrianPolicy}
	u^{\text{ped}}_{t} = \left\{\begin{array}{ll} \text{walk}, & \text{if }  \text{TTC}_{t} \geq 3\si{\second}  \\%
		\text{wait},  & \text{otherwise }\end{array}\right..
\end{equation}
Once the pedestrian decides to take action $u^{\text{ped}}_{t}$ to \textit{walk}, the walking speed $v^{\text{ped}}_{\text{walk}}$ is kept until the pedestrian's goal state is reached. Note that the agent will also start walking when the \gls*{av} has passed the crossing by 4m.

\subsubsection{Level-2}
\label{subsubsec:level3ped}
The \textit{learning} pedestrian is a \gls*{drl} agent based on a \gls*{dqn} (see \autoref{subsec:DQN}). The pedestrian's state $s_t^{\text{ped}}$ at time $t$ is defined as the vector
\begin{equation}
	\label{eqn:StatePed}
	s_t^{\text{ped}} = 
	    \begin{array}{ll}
	    	\Big[\text{TTC}_t,\, |{{v}}^{\text{ped}}_{t}|,\, |{{v}}^{\text{ped}}_{\text{walk}}|,\, |{{v}}^{\text{AV}}_{t}|,\, |{{a}}^{\text{AV}}_{t}|,\, \\%
        	 {\Delta\boldsymbol{x}}^{\text{rel}}_{t},\, \text{PDTC}_t,\, b^{\text{street}},\, b^{\text{street}}_{\text{side}}
        	\Big]^{\top}
	    \end{array},
\end{equation}
with nine components described as follows:
\begin{itemize}
	\label{eqn:StateSpacePed}
	\item ${\text{TTC}_t} \in \mathbb{R}$ represents the current \gls*{ttc} value
	\item $|{{v}}^{\text{ped}}_{t}| \in \mathbb{R}^{+}$ is the pedestrian's current absolute velocity
	\item $|{v}^{\text{ped}}_{\text{walk}}| \in \mathbb{R}^{+}$ is the pedestrian's constant absolute walking speed once the agent decides to start walking
	\item $|{{v}}^{\text{AV}}_{t}|  \in \mathbb{R}^{+}$ describes the absolute velocity of the \gls*{av}
	\item $|{{a}}^{\text{AV}}_{t}|  \in \mathbb{R}^{+}$ is the absolute acceleration of the \gls*{av}
	\item ${\Delta\boldsymbol{x}}^{\text{rel}}_{t} \in \mathbb{R}^2$ measures the two-dimensional position of the pedestrian relative to the \gls*{av}'s center point
	\item $\text{PDTC}_t \in \mathbb{R}^{+}$ is defined as the remaining crossing distance for the pedestrian to reach its goal position
	\item $b^{\text{street}}  \in \mathbb{R}^{+}$ is the width of the street
	\item $b^{\text{street}}_{\text{side}} \in \{\text{left},\text{right}\}$ indicates from which street side the pedestrian will start crossing.
\end{itemize}

The state $s_t^{\text{ped}}$ is an element of the state space $\mathcal{S}^\text{ped}$, while the discrete actions space $\mathcal{U}^{\text{ped}}$ allows for two choices:
\begin{equation}
	\label{eqn:ActionSpacePed}
	\mathcal{U}^{\text{ped}} = \left\{\text{wait}, \text{walk} \right\}.
\end{equation}
When the pedestrian decides to walk at time $t$, its velocity ${{v}}^{\text{ped}}_{t+1}$ at the next time step is set to $v^{\text{ped}}_{\text{walk}}$. The pedestrian's reward function $\mathcal{R}^{\text{ped}}$ with the reward $r^{\text{ped}}_{t+1}$ is based on
\begin{equation}
	\label{eqn:RewardFunctionPedModel}
	r^{\text{ped}}_{t+1} = - \tau^{\text{ped}} - \left\{\begin{array}{ll} \beta^{\text{ped}}, & \text{if collision = True} \\ 0, & \text{otherwise}\end{array}\right..
\end{equation}
The first term $\tau^{\text{ped}}$ penalizes each time step taken; we set $\tau^{\text{ped}}=0.01$. If a collision occurs, a penalty of $\beta^{\text{ped}}=10$ is added. While it is important to keep a balance between the two penalty terms, the choice of absolute values is justified empirically with the aim to minimize training instabilities. In summary, the pedestrian's goal is to reach the other street side as quickly as possible without risking a collision.

\subsection{AV Models}
\label{subsec:AVModels}
\subsubsection{Level-1}
Derived from a best response analysis, the \gls*{av}'s best velocity $v^{\text{AV}}_{t, \text{best}}$ is calculated at time $t$ according to
\begin{equation}
	\label{eqn:BestResponseCar}
	v^{\text{AV}}_{t, \text{best}}=\left\{
	\begin{array}{ll} 
		\sfrac{d_{\text{low}}}{\Delta t^{\text{ped}}_{t}}, & \text{if pedestrian walks}   \\   v^{\text{AV}}_{\text{limit}}, & \text{otherwise }
	\end{array}
	\right.,
\end{equation}
with $\Delta t^{\text{ped}}_{t}$ as the pedestrian's theoretical crossing duration. The longitudinal distance from the \gls*{av}'s front to the pedestrian is given by $d_{\text{low}}$. As it is not possible to set $v^{\text{AV}}_{t,\text{best}}$ in the simulation directly due to the \gls*{av}'s dynamic model, an acceleration value with $a^{\text{AV}}_{t, \text{best}}$ is set to reach $v^{\text{AV}}_{t,\text{best}}$ in minimal time instead. The list of the possible, discrete acceleration values $a^{\text{AV}}_{t}$ is given by the action space $\mathcal{U}^{\text{AV}}$ with
\begin{equation}
	\label{eqn:ActionSpaceCar}
	\mathcal{U}^{\text{AV}} = \left\{
	-9.8,
	-5.8,
	-3.8,
	0,
	1,
	3
	\right\}\frac{\si{\meter}}{\text{s}^2}.
\end{equation}

\subsubsection{Level-2}
This model of the \gls*{av} uses the \gls*{dqn} algorithm to enable the agent to learn from interaction with its environment. The \gls*{av}'s state $s_t^{\text{AV}}$ at time $t$ is element of the state space $\mathcal{S}^\text{AV}$, and the components of $s_t^{\text{AV}}$ form a vector
\begin{equation}
	\label{eqn:StateCar}
	s_t^{\text{AV}} = 
	    \begin{array}{ll}
	    	\Big[\text{TTC}_t,\, |{{v}}^{\text{ped}}_{t}|,\, |{{v}}^{\text{ped}}_{\text{walk}}|,\, |{{v}}^{\text{AV}}_{t}|,\, |{{a}}^{\text{AV}}_{t}|,\, \\
        	 {\Delta\boldsymbol{x}}^{\text{rel}}_{t},\, \text{PDTC}_t,\, b^{\text{street}},\, b^{\text{street}}_{\text{side}}
        	\Big]^{\top}
	    \end{array}.
\end{equation}
See \autoref{subsubsec:level3ped} for a description of these components. The \gls*{av}'s action space $\mathcal{U}^{\text{AV}}$ is equivalent to (\ref{eqn:ActionSpaceCar}); its reward function $\mathcal{R}^{\text{AV}}$ with reward $r^{\text{AV}}_{t+1}$ is described by 
\begin{equation}
	\label{eqn:RewardFunctionCarModel}
	\begin{aligned}
		r^{\text{AV}}_{t+1} = & - \tau^{\text{AV}} - \left\{\begin{array}{ll} \beta^{\text{AV}}, & \text{if collision = True} \\ 0, & \text{otherwise}\end{array}\right. \\%
		& - \left\{\begin{array}{ll} \psi^{\text{AV}} , & \text{if } v^{\text{AV}}_{t} > v^{\text{AV}}_{\text{limit}}  \\ 0, & \text{otherwise}\end{array}\right..
	\end{aligned}
\end{equation}
At each time step, the constant penalty $\tau^{\text{AV}}=0.01$ is given; $\beta^{\text{AV}}=10$ is the collision penalty. The speed penalty $\psi^{\text{AV}}=0.05$ is subtracted when the \gls*{av} drives faster than the speed limit $v^{\text{AV}}_{\text{limit}}$. The intention in this context is that the \gls*{av} should learn to follow the traffic rules but the possibility to pass the speed limit should be given in emergency situations. 

\subsection{System Formulation}
\subsubsection{Partially observable Markov decision process}
\label{subsec:POMDB}
For the \gls*{drl} case in setting-1, we define a \gls*{pomdp} with tuple $(\mathcal{S}^{\text{AV}}, \mathcal{Z}^{\text{AV}}, \mathcal{U}^{\text{AV}}, \mathcal{T}, \mathcal{O}, \mathcal{R}^\text{AV}, \gamma)$ as follows:
\begin{itemize}
	\item States $s^{\text{AV}}_{t}$, see (\ref{eqn:StateCar}), are elements of a state space $\mathcal{S}^{\text{AV}}$.
	\item Due to the partial observability of the states, the \gls*{av} observes $z^{\text{AV}}_{t} \in \mathcal{Z}^{\text{AV}}$ which is described by the observation function $\mathcal{O}$ according to (\ref{eqn:ObsFunc}) instead of $s^{\text{AV}}_t$.
	\item $\mathcal{U}^{\text{AV}}$ is the action space of the \gls*{av} as introduced in (\ref{eqn:ActionSpaceCar}).
	\item $\mathcal{T}$ is the state transition function defined by the mapping $\mathcal{T}: \mathcal{S}^{\text{AV}} \times \mathcal{U}^{\text{AV}} \times \mathcal{S}^{\text{AV}} \rightarrow [0, 1]$ of the current state $s^{\text{AV}}_{t}$ to the probability of transitioning to the next state $s^{\text{AV}}_{t+1}$.
	\item Based on the reward function $\mathcal{R}^\text{AV}$ presented in (\ref{eqn:RewardFunctionCarModel}), the \gls*{av} receives the scalar reward $r^{\text{AV}}_{t+1}$.
	\item The discount factor $\gamma$ is used to weigh the importance of immediate to future rewards.
\end{itemize}

\subsubsection{Partially observable Markov game}
\label{subsec:POMG}
The system formulation as a \gls*{pomdp} is not sufficient for the \gls*{dmarl} approach in setting-2. Therefore, we introduce a \gls*{pomg} for the agents $\mathcal{W}=\{w^{\text{AV}}, w^{\text{ped}}\}$ which is described by the tuple $(\mathcal{W}, \mathcal{S}, \mathcal{Z}, \mathcal{U}, \mathcal{T},  \mathcal{O}, \mathcal{R}, \gamma)$ with following components:
\begin{itemize}
	\item The \gls*{av}'s states $s^{\text{AV}}_{t}$ are elements of the state space $\mathcal{S}^{\text{AV}}$ defined in (\ref{eqn:StateCar}), while the pedestrian's state space $\mathcal{S}^{\text{ped}}$ is presented in (\ref{eqn:StatePed}). The joint state $\boldsymbol{s}_t=(s^{{\text{AV}}}_{t}, s^{{\text{ped}}}_t)$ is element of the joint space $\mathcal{S}=\mathcal{S}^{\text{AV}} \times \mathcal{S}^{\text{ped}}$.
	
	\item The observation function $\mathcal{O}$, see (\ref{eqn:ObsFunc}), defines the observations of the \gls*{av} as $z^{\text{AV}}_{t}$ and $z^{\text{ped}}_{t}$ for the pedestrian.
	
	\item $\mathcal{U}^{\text{AV}}$ is the \gls*{av}'s action space given in (\ref{eqn:ActionSpaceCar}); see (\ref{eqn:ActionSpacePed}) for the pedestrian's action space $\mathcal{U}^{\text{ped}}$. The joint action $\boldsymbol{u}_t = (u^{\text{AV}}_{t}, u^{\text{ped}}_{t})$ with $\boldsymbol{u}_t \in \mathcal{U}$ is selected each step $t$.
	
	\item $\mathcal{T}$ is the joint state transition function with $\mathcal{T}: \mathcal{S} \times \mathcal{A} \times \mathcal{S}\rightarrow [0, 1]$ mapping the current joint state $\boldsymbol{s}_t$ to the probability of the next state $\boldsymbol{s}_{t+1}$ taking action $\boldsymbol{u}_t$.
	
	\item The \gls*{av}'s reward signal $r^{\text{AV}}_{t+1}$ follows $\mathcal{R}^{\text{AV}}$ given in (\ref{eqn:RewardFunctionCarModel}), and (\ref{eqn:RewardFunctionPedModel}) defines the pedestrian's reward function $\mathcal{R}^\text{ped}$.
	
	\item The same discount factor $\gamma$ is introduced for all agents.
\end{itemize}

\subsection{Deep Q-Networks}
\label{subsec:DQN}
All \gls*{drl} agents in this work are modeled by \glspl*{dqn}. As introduced in \cite{mnih2015human}, a \gls*{dqn} is an \textit{off-policy} method using a \gls*{nn} to learn the Q-function of an optimization problem iteratively. An enhanced version, called \gls*{ddqn} \cite{vanhasselt2016deep}, has the training target
\begin{equation}
	\label{eqn:DDQNTarget}
	y_t = \left\{\begin{array}{ll}
		 r_{t+1}, &\text {if goal state} \\
		 r_{t+1} + \gamma Q(s_{t+1}, u_{t+1} \mid \boldsymbol{\theta}^{-}_{i} ), &\text{otherwise}
	\end{array}\right.,
\end{equation}
which decouples the action selection and evaluation. $\boldsymbol{\theta}^{-}_{i}$ are the trainable parameters at training step $i$ of the target network, $\boldsymbol{\theta}_{i}$ the parameters of the decision network, respectively. The action $u_{t+1}$ of the next time step in (\ref{eqn:DDQNTarget}) is given through
\begin{equation}
	\label{eqn:utp1}
    u_{t+1}=\argmax_{u_{t+1}} Q(s_{t+1},u_{t+1} \mid \boldsymbol{\theta}_{i}).
\end{equation}
A sampling-based strategy is used in practice to estimate the error between the bootstrapped target $y_t$ and the prediction $Q\left(s_t, u_t \mid \boldsymbol{\theta}_{i}\right)$ over a batch of $M$ training samples. The experience replay buffer $\mathcal{E}$ is a data container storing the last $L$ experience $e_t=\left(s_t, u_t, r_{t+1}, s_{t+1}\right)$. To update $\boldsymbol{\theta}_{i}$, a batch of $M$ experiences are randomly sampled from $\mathcal{E}$, the loss between predictions and targets $y_t$ calculated, and $\boldsymbol{\theta}_{i+1}$ updated by gradient descent. For inference, the agent's behavioral policy is obtained by means of the greedy policy
\begin{equation}
	u_t=\argmax_{u_t}\,Q\left(s_t, u_t \mid \boldsymbol{\theta}_{i}\right),
\end{equation}
and an $\varepsilon$-greedy policy is used for exploration: At each time step $t$, a random action $u_t$ is chosen with probability $\varepsilon$; we decrease $\varepsilon$ exponentially. Early results in this work have shown that the \gls*{av}'s unbalanced action space, i.e., the deceleration values are higher than the acceleration values (causing the \gls*{av} to stand still in expectation with initially uniformly random actions after a finite number of time steps), introduces instability into the learning process. We overcome this issue by moving the probability mass so that acceleration values are selected with higher probability. In our implementation, we improve the \gls*{ddqn} \cite{vanhasselt2016deep} method by using the combined replay buffer \cite{zhang2017deep}, multi-step learning \cite{sutton1988learning}, and dueling heads \cite{wang2016dueling} extensions. Additionally, gradients larger than 10 are clipped for numerical stability, and the Huber loss is used with a linear slope starting at $\delta^{\text{Huber}}=1$ to calculate the error. The \gls*{dqn} is updated each time step a new experience is generated in our simulation. 

The \textit{independent learning} scheme is used for implementation of the \gls*{dmarl} case: From the perspective of a single agent, the other agent is assumed to be part of the environment, allowing the use of the single-agent \gls*{dqn} method as introduced earlier. This approach harms the assumption of a stationary environment; successful implementation in other works and simplicity motivate the use here.

\subsection{Simulation Setup}
The initial longitudinal velocity of the \gls*{av} $v^{\text{AV}}_{\text{init}}$ is sampled from a uniform distribution $v^{\text{AV}}_{\text{init}} \sim \mathcal{U}(30\frac{\si{\km}}{\si{\hour}}, 50\,\frac{\si{\km}}{\si{\hour}})$, reflecting the typical velocities driven in urban areas in Germany. The initial \gls*{ttc} value is randomly sampled with $\text{TTC}_{\text{init}} \sim \mathcal{U}(1.0\si{\s},\, 5.0\si{\s})$, defining the \gls*{av}'s initial position $\boldsymbol{x}^{\text{AV}}_{\text{init}}$ with a distance to the crossing of $\text{TTC}_{\text{init}} \cdot v^{\text{AV}}_{\text{init}}$. The goal state of the \gls*{av} $\boldsymbol{x}^{\text{AV}}_{\text{goal}}$ is reached when the \gls*{av} has passed the crosswalk by $\zeta^{\text{AV}}=10\si{\meter}$. We take into account the German speed limit of  $v^{\text{AV}}_{\text{limit}} = 50\frac{\si{\km}}{\si{\hour}}$ in urban areas. The pedestrian's initial street side position $\boldsymbol{x}^{\text{ped}}_{\text{init}}$ depends on a random variable $b_{\text{side}}^{\text{street}}$ drawn uniformly from $\{\text{left}, \text{right}\}$. The variability of pedestrians' walking speed is reflected by selecting $v^{\text{ped}}_{\text{walk}}$ uniformly from $v^{\text{ped}}_{\text{walk}} \in \left\{1.16,\,1.38,\, 1.47,\,1.53,\,1.55\,\right\}\frac{\si{\meter}}{\si{\s}}$, representing typical pedestrian walking speeds \cite{willis2004human}. A value of $0.5\si{\m}$ is used for the safety margin $\zeta^{\text{ped}}$. Additional variability in the environment is introduced by selecting the street width $b^{\text{street}}$ uniformly from $\{6.0, 7.5\}\si{\meter}$.

A collision occurs when the inequality (\ref{eqn:collision}) is fulfilled, we set the collision margin to $\eta =0.5 \si{\meter}$. Analysis of early results showed that increasing the safety zone to $\eta_{\text{train}}=1.5 \si{\meter}$ during training leads to a reduction of the collision rate during evaluation. 
We use this strategy for the training of all level-2 agents. 
A good trade-off between a reasonable fast control input frequency and high computational costs is achieved by setting the time constant to $\delta_t=0.1\si{\second}$. Further computational costs associated with simulating potentially non-terminating training episodes are avoided by setting a timeout $\nu=15\si{\second}$ for the maximal episode duration. %
\section{Results}
For investigation of our proposed pedestrian models, all agents learn over 8,000 episodes with 800 episodes used for exploration (the first 250 episodes are completely random). Regarding the architecture of the \glspl*{dqn}, a fully-connected \gls*{nn} and a replay buffer size of 50,000 experiences is selected; hyperparameters were obtained after a limited parameter search. We use the exact same training settings for the \gls*{dmarl} approach of setting-2 for both agents. When we present results, the median value of 8 complete, independent training runs with different seeds are reported. The deviation between runs is indicated in form of an 80\%-confidence interval spanning from the 10\%-quantile to the 90\%-quantile.
	
\subsection{Performance Evaluation}
The effect of uncertainty in form of measurement noise, see (\ref{eqn:ObsFunc}), is evaluated with a fixed noise level $\alpha^{\text{AV}}=0.05$ for the AV and over $\alpha^{\text{ped}} \in \{0.0, 0.1, 0.2, 0.3, 0.4, 0.5\}$ for the pedestrian. We focus on the \gls*{av}'s response under uncertain pedestrian behavior as the pedestrian's estimations of its environment are unreliable, i.e., the pedestrian is likely to make a wrong crossing decision.

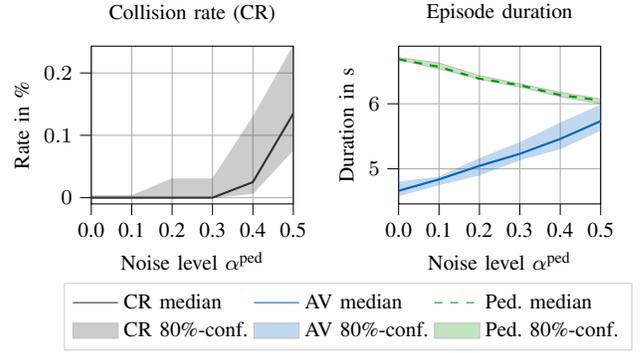
\begin{figure}
	\centering%
	\begin{tikzpicture}[scale=1]

	\definecolor{color0}{rgb}{0,0.396,0.741}
	\definecolor{color1}{rgb}{0.2, 0.2, 0.2}
	\definecolor{color2}{rgb}{0.0, 0.6, 0.0}

	\footnotesize

	\begin{groupplot}[group style={group size=2 by 1, horizontal sep=1.4cm, vertical sep=2cm}, width=0.24\textwidth]
		\nextgroupplot[
		legend cell align={left},
		legend style={fill opacity=0.8, draw opacity=1, text opacity=1, draw=white!80!black, align=left, font=\footnotesize},
		tick align=outside,
		tick pos=left,
		title={Collision rate (CR)},
		x grid style={white!69.0196078431373!black},
		xlabel={Noise level $\alpha^{\text{ped}}$},
		xmajorgrids,
		xmin=0, xmax=0.5,
		xtick style={color=black},
		xtick={0,0.1,0.2,0.3,0.4,0.5},
		xticklabels={0.0,0.1,0.2,0.3,0.4,0.5},
		y grid style={white!69.0196078431373!black},
		ylabel={Rate in \%},
		ymajorgrids,
		ymin=-0.01, ymax=0.243,
		ytick style={color=black},
		yticklabel style={
			/pgf/number format/fixed,
			/pgf/number format/precision=2},
		legend style={at={(0.5, -0.5)},anchor=north}
		]
		
		\path [draw=color1, fill=color1, opacity=0.25]
		(axis cs:0,0.003)
		--(axis cs:0,0)
		--(axis cs:0.1,0)
		--(axis cs:0.2,0)
		--(axis cs:0.3,0)
		--(axis cs:0.4,0.007)
		--(axis cs:0.5,0.077)
		--(axis cs:0.5,0.243)
		--(axis cs:0.5,0.243)
		--(axis cs:0.4,0.129)
		--(axis cs:0.3,0.03)
		--(axis cs:0.2,0.03)
		--(axis cs:0.1,0.003)
		--(axis cs:0,0.003)
		--cycle;

		\addplot [thick, color1, opacity=1]
		table {%
			0 0
			0.1 0
			0.2 0
			0.3 0
			0.4 0.025
			0.5 0.135
		};
		
	    \nextgroupplot[
		legend cell align={left},
		legend style={
			fill opacity=0.8,
			draw opacity=1,
			text opacity=1,
		    at={(-0.25, -0.5)},
			anchor=north,
			draw=white!80!black,
			font=\footnotesize
		},
	    legend columns=3,
	    tick align=outside,
		tick pos=left,
		title={Episode duration},
		x grid style={white!69.0196078431373!black},
		xlabel={Noise level $\alpha^{\text{ped}}$},
		xmajorgrids,
		xmin=0, xmax=0.5,
		xtick style={color=black},
		xtick={0,0.1,0.2,0.3,0.4,0.5},
		xticklabels={0.0,0.1,0.2,0.3,0.4,0.5},
		y grid style={white!69.0196078431373!black},
		ylabel={Duration in s},
		ymajorgrids,
		ytick style={color=black}
		]
		
		\addlegendentry{CR median}
		\addlegendimage{line legend, draw=color1}
		
		\addlegendentry{AV median}
		\addlegendimage{line legend, draw=color0}
		
		\addlegendentry{Ped. median}
	    \addlegendimage{line legend, draw=color2, dashed}
	    
	    \addlegendentry{CR 80\%-conf.}
	    \addlegendimage{area legend, draw=black, fill=color1, opacity=0.25}
	    
	    \addlegendentry{AV 80\%-conf.}
		\addlegendimage{area legend, draw=black, fill=color0, opacity=0.25}
		
		\addlegendentry{Ped. 80\%-conf.}
		\addlegendimage{area legend, draw=black, fill=color2, opacity=0.25}

		\path [draw=color0, fill=color0, opacity=0.25]
		(axis cs:0,4.795534)
		--(axis cs:0,4.585973)
		--(axis cs:0.1,4.756218)
		--(axis cs:0.2,4.906855)
		--(axis cs:0.3,5.13804781974592)
		--(axis cs:0.4,5.30966769736567)
		--(axis cs:0.5,5.59315307047164)
		--(axis cs:0.5,5.97806936788488)
		--(axis cs:0.5,5.97806936788488)
		--(axis cs:0.4,5.70248261977325)
		--(axis cs:0.3,5.402299)
		--(axis cs:0.2,5.157422)
		--(axis cs:0.1,4.87271303630363)
		--(axis cs:0,4.795534)
		--cycle;
		
		\path [draw=black, fill=color2, opacity=0.25]
		(axis cs:0,6.71097999999999)
		--(axis cs:0,6.67465599999999)
		--(axis cs:0.1,6.53965964736473)
		--(axis cs:0.2,6.37876974412323)
		--(axis cs:0.3,6.25898645253575)
		--(axis cs:0.4,6.10904000364021)
		--(axis cs:0.5,6.00111884047158)
		--(axis cs:0.5,6.07399664017874)
		--(axis cs:0.5,6.07399664017874)
		--(axis cs:0.4,6.17724378779146)
		--(axis cs:0.3,6.30689049554866)
		--(axis cs:0.2,6.43244299999999)
		--(axis cs:0.1,6.62618599999999)
		--(axis cs:0,6.71097999999999)
		--cycle;
		
		\addplot [thick, color0, opacity=1]
		table {%
			0 4.662895
			0.1 4.83827
			0.2 5.04584375312593
			0.3 5.23096510153045
			0.4 5.46013224322432
			0.5 5.73320049748808
		};
		
		\addplot [thick, color2, dashed, opacity=1]
		table {%
			0 6.68345901490148
			0.1 6.56722999999999
			0.2 6.38497499999999
			0.3 6.28661999999999
			0.4 6.13412037761328
			0.5 6.05149510838831
		};

	\end{groupplot}
	
\end{tikzpicture}%
	\caption{Results of setting-1: \gls*{drl}-based \gls*{pcam} system with a rational pedestrian model.}%
	\label{fig:Setting1Results}
\end{figure}

\autoref{fig:Setting1Results} visualizes the performance of the \gls*{drl}-based \gls*{pcam} system in setting-1. It can be seen that the \gls*{av} learns a flawless behavior without any collision when the pedestrian behavior is certain, i.e., the collision rate at $\alpha^{\text{AV}}=0.0$ and $\alpha^{\text{AV}}=0.1$ is $0.0\%$. A higher degree of uncertainty increases the collision rate as expected. Remarkably, the \gls*{av} is still able to mitigate most collisions when the pedestrian acts nearly unpredictably at $\alpha^{\text{AV}}=0.5$ with a $0.135\%$ collision rate. Increasing the degree of noise makes the pedestrian more likely to cross leading to a reduced episode duration for the pedestrian. The \gls*{av} accounts for this difficult-to-predict pedestrian behavior; its episode duration increases by approximately 23\% from 4.663s to 5.722s.

\begin{figure}
	\centering%
	\begin{tikzpicture}[scale=1]

	\definecolor{color0}{rgb}{0,0.396,0.741}
	\definecolor{color1}{rgb}{0.2, 0.2, 0.2}
	\definecolor{color2}{rgb}{0.0, 0.6, 0.0}

    \footnotesize
	\begin{groupplot}[group style={group size=2 by 1, horizontal sep=1.4cm, vertical sep=2cm}, width=0.24\textwidth]
		\nextgroupplot[
		legend cell align={left},
		legend style={fill opacity=0.8, draw opacity=1, text opacity=1, draw=white!80!black, align=left, font=\footnotesize},
		tick align=outside,
		tick pos=left,
		title={Collision rate (CR)},
		x grid style={white!69.0196078431373!black},
		xlabel={Noise level $\alpha^{\text{ped}}$},
		xmajorgrids,
		xmin=0, xmax=0.5,
		xtick style={color=black},
		xtick={0,0.1,0.2,0.3,0.4,0.5},
		xticklabels={0.0,0.1,0.2,0.3,0.4,0.5},
		y grid style={white!69.0196078431373!black},
		ylabel={Rate in \%},
		ymajorgrids,
		ymin=-0.01, ymax=0.243,
		ytick style={color=black},
		yticklabel style={
			/pgf/number format/fixed,
			/pgf/number format/precision=2},
		legend style={at={(0.5, -0.5)},anchor=north}
		]
		
	    \path [draw=color1, fill=color1, opacity=0.25]
		(axis cs:0,0.094)
		--(axis cs:0,0)
		--(axis cs:0.1,0)
		--(axis cs:0.2,0)
		--(axis cs:0.3,0.02)
		--(axis cs:0.4,0.007)
		--(axis cs:0.5,0)
		--(axis cs:0.5,0.166)
		--(axis cs:0.5,0.166)
		--(axis cs:0.4,0.304)
		--(axis cs:0.3,0.264)
		--(axis cs:0.2,0.083)
		--(axis cs:0.1,0.051)
		--(axis cs:0,0.094)
		--cycle;
		
		\addplot [thick, color1, opacity=1]
		table {%
			0 0
			0.1 0
			0.2 0.02
			0.3 0.125
			0.4 0.11
			0.5 0.03
		};

	    \nextgroupplot[
		legend cell align={left},
		legend style={
			fill opacity=0.8,
			draw opacity=1,
			text opacity=1,
		    at={(-0.25, -0.5)},
			anchor=north,
			draw=white!80!black,
			font=\footnotesize
		},
	    legend columns=3,
	    tick align=outside,
		tick pos=left,
		title={Episode duration},
		x grid style={white!69.0196078431373!black},
		xlabel={Noise level $\alpha^{\text{ped}}$},
		xmajorgrids,
		xmin=0, xmax=0.5,
		xtick style={color=black},
		xtick={0,0.1,0.2,0.3,0.4,0.5},
		xticklabels={0.0,0.1,0.2,0.3,0.4,0.5},
		y grid style={white!69.0196078431373!black},
		ylabel={Duration in s},
		ymajorgrids,
		ytick style={color=black}
		]
		
		\addlegendentry{CR median}
		\addlegendimage{line legend, draw=color1}
		
		\addlegendentry{AV median}
		\addlegendimage{line legend, draw=color0}
		
		\addlegendentry{Ped. median}
	    \addlegendimage{line legend, draw=color2, dashed}
	    
	    \addlegendentry{CR 80\%-conf.}
	    \addlegendimage{area legend, draw=black, fill=color1, opacity=0.25}
	    
	    \addlegendentry{AV 80\%-conf.}
		\addlegendimage{area legend, draw=black, fill=color0, opacity=0.25}
		
		\addlegendentry{Ped. 80\%-conf.}
		\addlegendimage{area legend, draw=black, fill=color2, opacity=0.25}
		
	\path [draw=color0, fill=color0, opacity=0.25]
		(axis cs:0,5.16584820882088)
		--(axis cs:0,4.579899)
		--(axis cs:0.1,4.45690776012804)
		--(axis cs:0.2,4.8294042253352)
		--(axis cs:0.3,5.62458441117195)
		--(axis cs:0.4,5.89775103669625)
		--(axis cs:0.5,6.16536299999999)
		--(axis cs:0.5,6.5147878976629)
		--(axis cs:0.5,6.5147878976629)
		--(axis cs:0.4,6.43993168051178)
		--(axis cs:0.3,6.22267806363514)
		--(axis cs:0.2,5.99697150850448)
		--(axis cs:0.1,5.78556499999999)
		--(axis cs:0,5.16584820882088)
		--cycle;
		
		\path [draw=black, fill=color2, opacity=0.25]
		(axis cs:0,7.00816699999999)
		--(axis cs:0,6.39175871047104)
		--(axis cs:0.1,5.883378)
		--(axis cs:0.2,5.74309356936946)
		--(axis cs:0.3,5.53636183737099)
		--(axis cs:0.4,5.53664871730695)
		--(axis cs:0.5,5.51093983169705)
		--(axis cs:0.5,5.68731698932399)
		--(axis cs:0.5,5.68731698932399)
		--(axis cs:0.4,5.59438591734295)
		--(axis cs:0.3,5.96087021249939)
		--(axis cs:0.2,7.00690227256353)
		--(axis cs:0.1,7.01563399999999)
		--(axis cs:0,7.00816699999999)
		--cycle;

		\addplot [thick, color0, opacity=1]
		table {%
			0 4.77695790282439
			0.1 5.01778664164164
			0.2 5.3991150040032
			0.3 6.00726362756669
			0.4 6.02542817063096
			0.5 6.34557282722566
		};

		\addplot [thick, color2, opacity=1, dashed]
		table {%
			0 6.65295499999999
			0.1 6.35344798298298
			0.2 6.04873561012202
			0.3 5.75800421884156
			0.4 5.55185665940281
			0.5 5.56006686217595
		};
		
	\end{groupplot}
	
\end{tikzpicture}%
	\caption{Results of setting-2: \gls*{dmarl}-based \gls*{pcam} system with a learning pedestrian model and learning AV, both using \gls*{drl}.}%
	\label{fig:Setting2Results}
\end{figure}
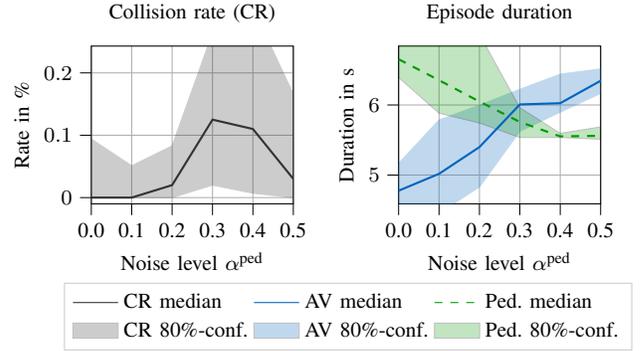

The results of the novel \gls*{dmarl} approach with the learning \gls*{av} and pedestrian models are presented in \autoref{fig:Setting2Results}. It is evident that the agents learn in presence of low uncertainty, i.e., $\alpha^{\text{ped}}=0.0$ and $\alpha^{\text{ped}}=0.1$, to avoid collisions completely. The highest median collision rate is obtained at $\alpha^{\text{ped}}=0.3$ with a rate of 0.125\%; the highest upper bound of the collision rate's confidence interval occurs at $\alpha^{\text{ped}}=0.4$ with 0.304\%. From a noise level of $\alpha^{\text{ped}}=0.4$ and higher, the agents achieve a reduction in collision rate again despite the higher uncertainty. Analysis of the agents' behavior shows that this effect is due to the pedestrian who tends to cross the street recklessly, expecting the \gls*{av} to react to avoid a collision. This situation correlates with the \textit{crosswalk chicken} problem mentioned in \cite{millard2018pedestrians} as the pedestrian learns to dominate the \gls*{av}'s strategy. From the \gls*{av}'s perspective, it is easier to adapt to this quasi-deterministic policy instead of the pedestrian's strategies learned at lower noise levels.

\begin{figure}
	\centering%
	\begin{tikzpicture}[scale=1]
	\definecolor{color0}{rgb}{0,0.396,0.741}
	\definecolor{color1}{rgb}{0.2, 0.2, 0.2}
	\definecolor{color2}{rgb}{0.0, 0.6, 0.0}

    \usetikzlibrary{patterns}
	\footnotesize

	\begin{axis}[
		width  = 0.5\textwidth,          
		height = 0.24\textwidth,
		title ={Collision rate (CR)},
		major x tick style = transparent,
		bar width=0.3cm,
		ybar=1.2\pgflinewidth,
		ymajorgrids = true,
		xlabel={Noise level $\alpha^{\text{ped}}$},
		xtick = data,
		scaled y ticks = false,
		enlarge x limits=0.10,
		ylabel={Rate in \%},
		legend cell align=left,
		legend style={
			fill opacity=0.8,
			draw opacity=1,
			text opacity=1,
		    at={(0.5, 0.985)},
			anchor=north,
			draw=white!80!black},
		legend columns=3,
		legend image code/.code={
        \draw [#1] (0cm,-0.1cm) rectangle (0.2cm,0.15cm); },
		]
		
		\addplot+[ybar, style={draw=black, fill=color1, opacity=0.5}, error bars/.cd, y dir=both, y explicit, error bar style = {color1, opacity=1}]
		coordinates {
			(0.0, 0) += (0,0.003) -= (0,0.0)
			(0.1, 0.0) += (0,0.003) -= (0,0.000)
			(0.2, 0.0) += (0,0.03) -= (0,0.0)
			(0.3, 0.0) += (0,0.03) -= (0,0.00)
			(0.4, 0.025) += (0,0.104) -= (0,0.018)
			(0.5, 0.135) += (0,0.108) -= (0,0.058)};
	
		\addplot+[ybar, style={draw=black, fill=color0, opacity=0.5}, error bars/.cd, y dir=both, y explicit,  error bar style = {color0, opacity=1}]
		coordinates {
			(0.0, 0) += (0,0.094) -= (0,0.0)
			(0.1, 0.0) += (0,0.051) -= (0,0.000)
			(0.2, 0.02) += (0,0.081) -= (0,0.0)
			(0.3, 0.125) += (0,0.139) -= (0,0.123)
			(0.4, 0.11) += (0,0.194) -= (0,0.103)
			(0.5, 0.03) += (0,0.136) -= (0,0.03)};
		
		\addplot+[ybar, draw=black, fill=color2, opacity=0.5]
		coordinates {
			(0.0, 0)
			(0.1, 0.0) 
			(0.2, 0.015)
			(0.3, 0.04) 
			(0.4, 0.155)
			(0.5, 0.72)};
		
		\legend{Setting-1, Setting-2, Setting-X}
		
	\end{axis}
\end{tikzpicture}%
	\caption{Median collision rate in all settings with 80\%-conf. error bars.}%
	\label{fig:collisionComparison}
\end{figure}
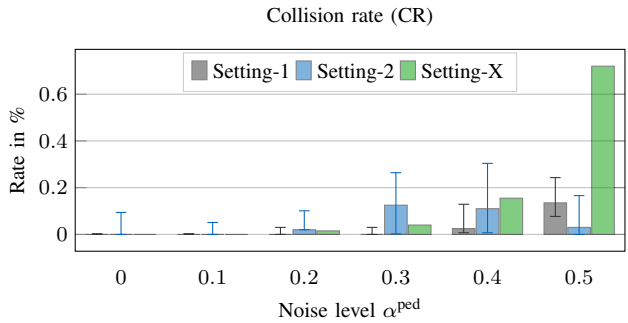

For performance comparison, \autoref{fig:collisionComparison} presents the collision rates under uncertainty for the three introduced settings. First, all agents learn to avoid collisions without mistakes for $\alpha^{\text{ped}}=0.0$ and $\alpha^{\text{ped}}=0.1$. While the benchmark case of setting-X shows comparable results, the \gls*{drl} and \gls*{dmarl}-based settings outperform it for the two highest uncertainty levels. It can be reasoned that these agents are able to learn a noise model implicitly, thereby following a more conservative but safer strategy. Note that the best model at $\alpha^{\text{ped}}=0.5$ is the \gls*{pcam} system using \gls*{dmarl} but otherwise the \gls*{drl}-based approach of setting-1 is superior with respect to the collision rates.
The authors assume that applied to the real world, the \gls*{dmarl} approach would outperform the single-agent \gls*{pcam} system as the policy is learned in response to a wider range of behaviors.

\subsection{Behavior Analysis of Setting-2}
\begin{figure}[!htbp]%
	\centering%
    \includegraphics[width=0.49\textwidth]{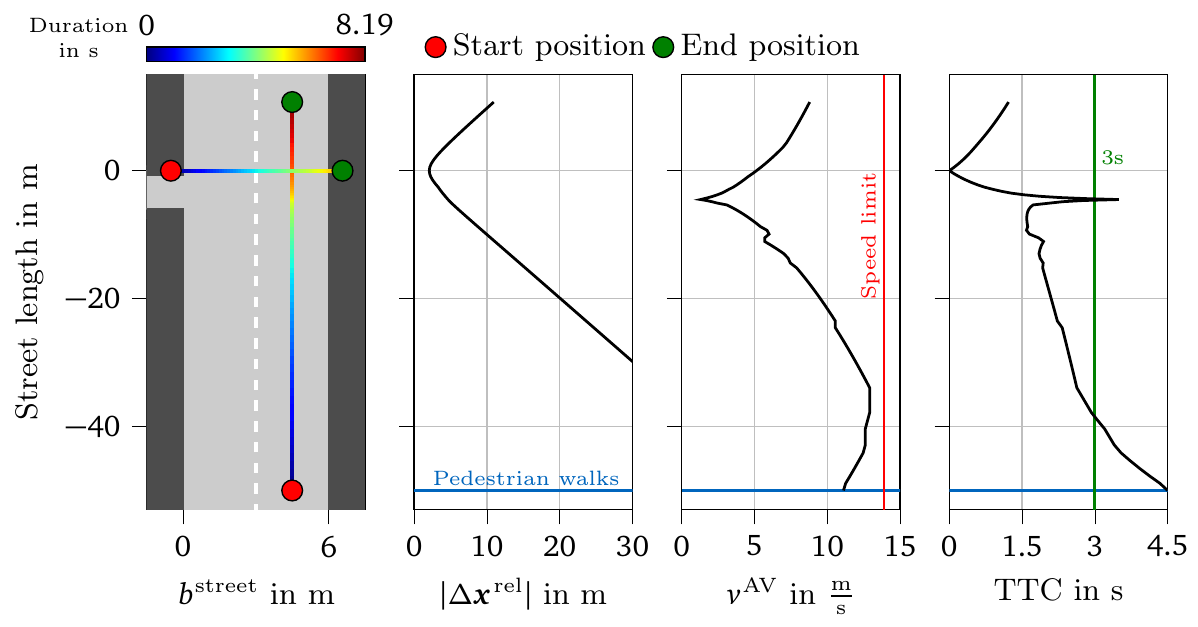}
    \caption{In this exemplary behavior of setting-2, the learning \gls*{av} model accelerates slightly at the episode start since the scenario is initialized with a high \gls*{ttc} value of 4.5s. At $\text{TTC}<3\si{\second}$, the \gls*{av} starts to reduce its velocity continuously to a near standstill when the pedestrian is directly in front. Interestingly, this TTC value corresponds to a real-world pedestrian's decision threshold as found in \cite{schmidt2009pedestrians}. As soon as the potential collision has been avoided, the \gls*{av} accelerates again to minimize the episode duration.}%
	\label{fig:setting2behavior}
\end{figure}

As demonstrated in \autoref{fig:setting2behavior}, the \gls*{av} model learns a policy that reduces the episode duration efficiently while keeping the speed limit and minimizing collisions. Although a real-world validation should be conducted in the future, the authors suspect that there are several situations in which the learning pedestrian model exhibits similarities to real human behavior, e.g., the pedestrian starts walking immediately in cases of a high initial distance of over 45m (see \autoref{fig:setting2behavior}). Another discovery is that the pedestrian's crossing decision is less reckless when starting from the left street side as the distance to the \gls*{av}'s lane is greater leading to a higher risk.

\section{Conclusion}
We have presented a new \gls*{pcam} system for \glspl*{av}, introducing a novel \gls*{dmarl}-based approach to model the vehicle-pedestrian interaction at crosswalks and analyzing the influence of observation uncertainty on the decision-making of the agents. Results show that while the \gls*{drl}-based approach paired with a deterministic pedestrian model achieves reliable performance over a large spectrum of uncertainty levels, the system using \gls*{dmarl} is exposed to a larger diversity of pedestrian behaviors retaining reliable collision avoidance even under uncertain pedestrian behavior. Subsequent works should validate similarities of the learned pedestrian behavior to real human behavior; our initial analysis indicates similar characteristics. To improve the proposed \gls*{pcam} system further, a more complex simulator (e.g., CARLA) should be used while extending the scenario to multiple road users. It may also be of interest to improve the independent \gls*{dmarl} training scheme in the future.
\vspace*{\fill}

\bibliographystyle{IEEEtran}
\bibliography{references}

\end{document}